# Gender Bias in Large Language Models for Healthcare: Assignment Consistency and Clinical Implications

Mingxuan Liu[1,2], Yuhe Ke[2,3,4], Wentao Zhu[1,2], Mayli Mertens[5,6], Yilin Ning[1,2], Jingchi Liao[1,2], Chuan Hong[7], Daniel Shu Wei Ting[1,8,9], Yifan Peng[10,11], Danielle S. Bitterman[12], Marcus Eng Hock Ong[1,13,14], Nan Liu[1,2,7,13,15]*

[1] Centre for Quantitative Medicine, Duke-NUS Medical School, Singapore, Singapore

[2] Duke-NUS AI + Medical Sciences Initiative, Duke-NUS Medical School, Singapore, Singapore

[3] Department of Anaesthesiology and Perioperative Medicine, Singapore General Hospital, Singapore

[4] Data Science and Artificial Intelligence Lab, Singapore General Hospital, Singapore

[5] Centre for Ethics, Department of Philosophy, University of Antwerp, Antwerp, Belgium

[6] Antwerp Center on Responsible AI, University of Antwerp, Antwerp, Belgium

[7] Department of Biostatistics and Bioinformatics, Duke University School of Medicine, Durham, NC, USA

[8] Artificial Intelligence Office, SingHealth, Singapore, Singapore

[9] Singapore National Eye Centre, Singapore, Singapore

[10] Department of Population Health Sciences, Weill Cornell Medicine, New York, NY, USA

[11] Department of Radiology, Weill Cornell Medicine, New York, NY, USA

[12] Artificial Intelligence in Medicine Program, Mass General Brigham, Harvard Medical School, Boston, MA, USA

[13] Pre-hospital & Emergency Research Centre, Health Services Research & Population Health, Duke-NUS Medical School, Singapore, Singapore

[14] Department of Emergency Medicine, Singapore General Hospital, Singapore

[15] NUS Artificial Intelligence Institute, National University of Singapore, Singapore, Singapore

*Correspondence: Nan Liu, Centre for Quantitative Medicine, Duke-NUS Medical School, 8 College Road, Singapore 169857, Singapore

Email: liu.nan@duke-nus.edu.sg

**Abstract**

The integration of large language models (LLMs) into healthcare holds promise to enhance clinical decision-making, yet their susceptibility to biases remains a critical concern. Gender has long influenced physician behaviors and patient outcomes, raising concerns that LLMs assuming human-like roles—such as clinicians or medical educators—may replicate or amplify gender-related biases. Using case studies from the New England Journal of Medicine Challenge (NEJM), we assigned genders (female, male, or unspecified) to multiple open-source and proprietary LLMs. We evaluated their response consistency across LLM-gender assignments regarding both LLM-based diagnosis and models' judgments on the clinical relevance or necessity of patient gender. In our findings, diagnoses were relatively consistent across LLM genders for most models. However, for patient gender's relevance and necessity in LLM-based diagnosis, all models demonstrated substantial inconsistency across LLM genders, particularly for relevance judgements. Some models even displayed a systematic female-male disparity in their interpretation of patient gender. These findings present an underexplored bias that could undermine the reliability of LLMs in clinical practice, underscoring the need for routine checks of identity-assignment consistency when interacting with LLMs to ensure reliable and equitable AI-supported clinical care.

**Keywords**

AI fairness, fairness in healthcare, large language model (LLM), bias in LLM

# Introduction

Artificial intelligence (AI), particularly large language models (LLMs) such as ChatGPT[1], has gained significant attention in healthcare[2] and is being rapidly applied to various clinical domains, including anesthesiology[3], dermatology[4], mental health[5], and medical education[6]. Physicians and patients are actively turning to LLMs as accessible tools for clinical decision support and medical consulting.[7-10] As these tools become more integrated into clinical workflows and medical education settings[11], questions arise about how LLMs might reflect or amplify biases that have long been observed in human practice.

Gender has historically shaped clinical practice, influencing not only the behavior and decision-making of clinicians but also the presentation of patient symptoms and treatment outcomes. Decades of research have shown that patients' gender can affect diagnostic accuracy[12], referral patterns[13], and treatment decisions[14]. For example, compared with men experiencing similar symptoms, women presenting with chest pain are less likely to be referred for cardiology evaluation or to receive appropriate medications.[15] Beyond patients' gender, physicians' gender and implicit biases have also been shown to influence clinical judgments[16], communication styles[17], and, ultimately, patient outcomes[18,19].

As LLMs increasingly adopt human-like roles—such as virtual physicians or medical educators—the gender-related dynamics that shape human interactions may similarly surface in AI agents.[20-22] Prior studies have shown that LLMs can internalize social biases from their training data[7] and display reasoning errors reminiscent of human cognitive fallacies[23]. Most existing investigations have focused on how models respond to various clinical vignettes or perceive patient and physician demographics.[7,24-26] In these cases, the model acts as an observer or judge of others' genders. However, much less is known about how LLMs behave when they themselves are assigned a gender identity in professional roles (i.e., gendered personas), either explicitly (e.g., through prompts such as "You are a female/male doctor") or implicitly (e.g., via system defaults). Whether such gender assignments alter clinical responses remains an open question, and current bias evaluation frameworks largely overlook this dimension.[24,27]

We refer to the influence of gender assignment on model behavior as "model gender framing" or "LLM gender effect". Such framing can result in gender-preferential responses, even when the underlying clinical content remains unchanged, thereby compromising the consistency, neutrality, and fairness expected in medical AI outputs[28]. While stylistic personalization may be acceptable and even desirable in some settings (e.g., to enhance user engagement in mental health applications[5]), variation in diagnostic decisions or inconsistent judgments about the clinical importance of certain factors raises serious clinical and ethical concerns. Specifically, when a model's assigned gender directly alters how it interprets the importance of a patient's gender in diagnoses, it may inadvertently undermine fairness, reinforce stereotypes, and erode trust in AI-assisted healthcare systems.

In this study, we investigated whether assigning gendered personas to LLMs alters their clinical responses. Specifically, we assessed the consistency of models' diagnosis outputs and their judgments about the importance of patient gender in the LLM-based decision-making—whether deemed relevant or necessary—under different LLM-gender assignments. Using a high-fidelity diagnostic dataset from the New England Journal of Medicine (NEJM) Challenge, we evaluated both open-source and proprietary models. Our findings reveal that gender assignment can influence how LLMs weigh patient gender in diagnosis, introducing variability that may reinforce gender bias and stereotypes, particularly in open-source models. This observed inconsistency across gender assignments raises questions about the reliability and neutrality of LLMs and underscores the need to assess LLM behavior across a broader range of identity assignments—including but not limited to gender—to ensure fair, trustworthy, and equitable AI-assisted clinical decision-making.

## Methodology

### Dataset

We evaluated the LLMs using the NEJM Image Challenge data (https://www.nejm.org/image-challenge), an imaging quiz designed to test medical professionals' knowledge and diagnostic capabilities. Our dataset comprised challenge questions released between January 2020 and February 2024. Each challenge presents a real patient vignette, along with diagnostic images and five multiple-choice options, spanning specialties such as dermatology, cardiology, and infectious disease. To assess whether LLM gender effects vary by clinical context or question complexity, we stratified the cases by both specialty and difficulty, with difficulty estimated using the percentage of correct responses from NEJM subscribers.

For this dataset, Jin et al.[29] implemented the Generative Pre-trained Transformer 4 with Vision (GPT-4V) for image comprehension and provided expert-level evaluations for its diagnostic reasoning. To isolate the language capabilities of LLMs and minimize errors related to vision interpretation, we analyzed only the subset of cases where GPT-4V correctly identified all key visual features relevant to the diagnosis.

The final dataset consisted of 117 cases where GPT-4V successfully generated clinically sound image interpretations, as verified by human experts. These cases spanned eight medical specialties, with dermatology (22.2%) and gastroenterology (12.8%) being the most represented. All cases were classified as "easy" (59.8%) or "medium" (40.2%), as all "hard" cases were excluded due to GPT-4V's inability to successfully comprehend them. For patient gender, male-associated terms (e.g., "man", "boy", "male") appeared in 54.7% of cases, and female-associated terms (e.g., "woman", "girl", "female") appeared in 45.3% of cases.

**Models**

To evaluate consistency across assigned LLM genders, we selected a list of pre-trained foundational LLMs, including both open-source and proprietary models. Table 1 provides detailed model characteristics. The open-source models were Gemma-2-2B[30] (google/gemma-2-2b-it), Phi-4-mini[31] (microsoft/Phi-4-mini-instruct), and LLaMA-3.1-8B[32] (meta-llama/Llama-3.1-8B-Instruct). The proprietary models were OpenAI o3-mini[33], GPT-4.1[34], and Gemini 2.5 Pro[35]. Together, these models covered a wide parameter range, from lightweight (~ 2B) to heavyweight (> 200B). All models are safety-aligned, meaning they are fine-tuned to reject harmful queries and produce outputs that align with human values.[36]

Inference used two deployment strategies based on model availability. Proprietary models were accessed via vendor APIs: GPT-4.1 and OpenAI o3-mini through the OpenAI API, and Gemini 2.5 Pro via the Gemini API. Open-source models (Gemma-2B, Phi-4-mini, and LLaMA-3.1-8B) were implemented in the Google Colab on an NVIDIA A100 GPU using Hugging Face checkpoints.

**Prompt design and gender assignment**

As illustrated in Figure 1, for each clinical case, we assigned a gendered persona to each LLM (female, male, or unspecified). We then evaluated consistency across LLM genders in two dimensions: 1) the diagnosis and 2) the LLM's perceived importance of patient gender to that diagnosis.

To evaluate the importance of patient gender, we used two separate prompts per case. In the first prompt, the model was asked to provide a diagnosis, then judge whether patient gender is a relevant factor (i.e., "Yes" or "No"). In the second prompt, the model was asked to answer the same diagnostic question, but assess whether patient gender is necessary for diagnosis (i.e., "Yes" or "No"). Both prompts required the model to explain its assessment.

Each prompt included the case description, an expert-confirmed image interpretation[29], and a list of diagnosis options provided by NEJM. The models were instructed to select a single diagnosis from the list and assess the diagnostic importance of patient gender, providing a brief justification.

**Statistical analysis**

Our analysis examined two main metrics: diagnostic accuracy and consistency rate. Diagnostic accuracy was defined as the proportion of LLM-generated diagnoses that matched the ground-truth diagnoses. Consistency rate measured the proportion of cases in which outputs were identical across the three assigned LLM genders. This was calculated separately for each task: diagnosis, perceived relevance of patient gender, and perceived necessity of patient gender. For diagnosis, a case was considered consistent across genders if all three gender-assigned models provided either all correct or all incorrect diagnoses.

In addition, we calculated two further metrics: relevance and necessity rates, defined as the proportion of cases where a model answered "Yes" to whether patient gender was diagnostically relevant or necessary, respectively.

Bootstrapping was applied to estimate 95% confidence intervals for accuracy, relevance, and necessary rates. To ensure robustness, each prompt was executed three times per case with the same LLM gender to examine the effect of random variability in model outputs. To evaluate the significance of differences across assigned LLM genders, Cochran's Q tests were conducted to compare response distributions. All data analyses are performed in Python 3.9.7.

## Results

We evaluated how assignments of gendered personas influenced the behavior of LLMs in clinical diagnosis, focusing on the consistency regarding diagnostic accuracy as well as relevance and necessity judgments of patient gender in LLM-based diagnoses. Using a curated subset of the NEJM Image Challenge dataset[29], we analyzed both open-source and proprietary models across varying difficulty tiers and clinical specialties.

**Consistency across assigned LLM genders**

As shown in Figure 2, consistency rates across assigned LLM genders varied substantially across models and tasks. All models demonstrated satisfactory diagnostic consistency, with rates exceeding 90%. GPT-4.1 achieved the highest diagnostic consistency (97.44%), while Phi-4-mini recorded the lowest (91.45%). However, high consistency did not necessarily indicate high accuracy. For instance, Gemma-2B maintained high diagnostic consistency (>95%) but a relatively low average accuracy of 0.478, suggesting that a model can be consistently wrong across assigned LLM genders. In contrast, GPT-4.1 combined high consistency (>95%) with the highest average accuracy of 0.943.

Consistency rates were substantially lower when models assessed the clinical relevance of patient gender. Only GPT-4.1 achieved a consistency rate above 90%, whereas the proprietary model Gemini 2.5 Pro reached 81.2% and the open-source model LLaMA-3.1-8B produced the lowest rate at 58.97%. Necessity judgments were generally more stable than relevance judgments, with most models exceeding 95% consistency. Notably, for some models, such as the Gemini 2.5 Pro and Phi-4-mini, necessity consistency even surpassed diagnostic consistency. In contrast, LLaMA-3.1-8B (78.63%) and Gemma-2B (90.6%) showed lower necessity consistency than for diagnosis, but still higher than for relevance.

**Effects of LLM gender across difficulty tiers**

We assessed whether consistency across assigned LLM genders varied across question difficulty by analyzing relevance and necessity rates (i.e., the proportion of cases where models judged patient gender to be diagnostically relevant or necessary; see "Statistical analysis"). LLaMA-3.1-8B and GPT-4.1 were used as representative open-source and proprietary LLMs, respectively, as each achieved the highest diagnostic accuracy within its category.

As shown in Figure 3, for both easy and medium questions, inconsistencies across assigned LLM genders were greater when judging the relevance of patient gender than its necessity. LLaMA-3.1-8B, when assigned with a male persona, systematically showed higher relevance and necessity rates than when assigned with a female persona ($p < 0.001$ for both difficulty tiers). GPT-4.1 exhibited statistically significant differences in relevance rates across LLM genders only for easy questions ($p < 0.05$), with no significant differences for medium questions or for necessity. These findings indicate that judgments of patient gender's relevance are more vulnerable to variation across assigned LLM genders than necessity, with LLaMA-3.1-8B showing a persistent male–female disparity in interpreting the clinical importance of patient gender.

**Effects of LLM gender across clinical specialties**

Inconsistencies in judging the clinical importance of patient gender across assigned LLM genders were evident across multiple specialties (Figure 4). For LLaMA-3.1-8B, the model assigned with a female persona generally exhibited lower relevance and necessity rates than the model assigned with a male persona. For example, in cardiology, the relevance rate was 23.1% for the female-assigned model versus 30.8% for the male-assigned model, and the necessity rate was 0% versus 15.4%. Similar patterns were observed in dermatology, neurology, pulmonology, and other specialties, where the female-assigned model consistently yielded the lowest rates among the three assigned LLM genders. By contrast, GPT-4.1 showed smaller differences across LLM genders regarding patient gender's relevance. For instance, in neurology, GPT-4.1 reported a relevance rate of 16.7% across assigned LLM genders, whereas LLaMA-3.1-8B varied widely (0% female, 33.3% male, 50% unspecified). For necessity. GPT-4.1 also remained consistent across LLM genders in most specialties.

From a linguistic standpoint, cases in which patient gender is judged necessary should also be judged relevant, implying that necessity rates should not exceed relevance rates. However, this expected pattern was violated in several specialties. In infectious diseases, LLMs assigned as personas with unspecified gender showed higher necessity than relevance rates (LLaMA-3.1-8B: 40% for relevance and 60% for necessity; GPT-4.1: 20% for relevance and 40% for necessity). Similar misalignments were observed for LLaMA-3.1-8B when assigned a male persona in gastroenterology (13.3% relevance vs. 20.0% necessity) and pulmonology (22.2% relevance vs. 44.4% necessity).

**Discussion**

Our study highlights a critical yet underexplored dimension of bias in LLM: the impact of gender assignment on model behavior when LLMs are role-playing clinical professionals. While diagnoses could appear relatively consistent across LLM genders, all models demonstrated substantial inconsistency in assessing the importance of patient gender for diagnosis. These discrepancies persisted across medical specialties and difficulty levels, highlighting a systemic vulnerability that had not been previously documented. The fact that such variation can be triggered by a simple change in the model's characteristics (i.e., assigned LLM gender) underscores the urgent need to assess assignment consistency to ensure equitable and reliable AI behavior in clinical settings.

The same model can deliver inconsistent judgments when assigned different genders in role-playing clinical experts. This inconsistency becomes more pronounced for complex clinical questions (Figure 3) and persists across medical specialties (Figure 4), suggesting a general limitation rather than a domain-specific issue. This assigned LLM gender effect has several implications. First, it may reflect "token bias", in which the model relies on superficial cues (e.g., gender assignment) rather than genuinely understanding the underlying reasoning task.[37] Second, it points to a "systemic vulnerability", as gender-related judgments can be readily influenced through minor prompt manipulations, such as altering the assigned LLM genders. Third, such inconsistency raises concerns about the ability of LLMs in clinical reasoning, as it can both propagate biases and undermine trust in AI-assisted clinical tools.[38]

Variability in how LLMs interpret patient gender in diagnosis can shape how gender is discussed, perceived, and taught in medical contexts.[39] For example, given identical chest-pain symptoms in a middle-aged woman, a female-assigned model might emphasize her elevated sex-specific cardiac risk and recommend a full cardiac workup, whereas a male-assigned model might attribute those symptoms to stress and minimize further assessments. Such divergent recommendations—driven solely by the LLM genders—can reinforce stereotypes and exacerbate disparities[40]. Moreover, these inconsistencies may remain undetected in current bias evaluation paradigms, which often assume fixed model identities or overlook the assignment of identity altogether[24,27]. This oversight risks the silent accumulation of real-world harms and represents a missed opportunity to improve model consistency and advance equity.

Open-source models exhibited greater effects of LLM gender than proprietary models. While this may partly result from their smaller parameter sizes and lower overall capability, the high consistency but low diagnostic accuracy of Phi-4-mini—a small open-source model—shows that consistency and clinical utility can be decoupled. Transparency remains essential in healthcare, as it enables developers to continually improve models and empower professionals to hold them accountable.[41] Our findings underscore the need for rigorous bias audits and consistency checks across gender assignments. One potential solution is to enhance model alignment[42,43], which can involve constructing physician-validated datasets, fine-tuning models for consistent domain-specific reasoning[44], and systematically evaluating their behaviors across LLM gender assignments to build more trustworthy and equitable AI agents.

Balancing personalization with consistency is critical when designing and using LLMs in healthcare. Gendered LLM agents can improve user engagement, especially in patient-facing applications.[45,46] However, such personalization should not compromise clinical accuracy or introduce variability in interpreting sensitive information.[46] Our findings suggest that assigned gender roles may lead to inconsistent diagnostic interpretations and inadvertently reinforce stereotypes. As LLMs are increasingly integrated into clinical workflows[11], evaluations should explicitly assess the impact of identity assignments and implement safeguards to prevent these effects from undermining fairness and trust in clinical decision support.

This study has several limitations. First, we used only static, text-only data and did not incorporate longitudinal patient data or multimodal inputs. While this limits the scope to language-based reasoning, it allows for a more controlled analysis of gender-related variability in LLM behavior. Second, computational constraints prevented us from evaluating larger LLMs, which may exhibit different patterns from the models studied. Third, we did not assess the clinical appropriateness of including or excluding patients' gender for each specific diagnosis. Rather, we focused on assessing whether models' judgments varied under different LLM-gender assignments. In future work, we will integrate expert annotations to benchmark clinical justification and explore debiasing strategies such as counterfactual prompting to mitigate identity-assignment inconsistencies in LLM outputs.

## Conclusion

Our study shows that assigning gender to LLMs in their role as clinical professionals can induce inconsistencies in clinical reasoning. The same model may assess the diagnostic importance of patient gender differently, depending solely on the model's assigned genders.

Open-source models displayed both lower diagnostic accuracy and greater inconsistency across assigned LLM genders, compared with proprietary models. These results underscore the need to conduct routine assignment-consistency checks when deploying LLMs in healthcare and developing evaluation frameworks that address role-playing variability, thereby promoting more equitable and reliable AI-assisted clinical decision-making.

**Table 1.** Characteristics of Large language models involved in this study.

| Model | Type | Release Date | Model Size | Data Size | Safety alignment |
|---|---|---|---|---|---|
| Gemma-2-2B | Open-source | Jul 2024 | 2B | Various sources of 2 trillion tokens | Yes |
| Phi-4-mini | Open-source | Mar 2025 | 3.8B | Various sources of 6 trillion tokens | Yes |
| Llama-3.1-8B | Open-source | Jul 2024 | 13B | Various public available sources of ~15 trillion tokens | Yes |
| OpenAI o3-mini | Proprietary | Jan 2025 | ~200B | Not disclosed | Yes |
| GPT-4.1 | Proprietary | Apr 2025 | Undisclosed | Not disclosed | Yes |
| Gemini-2.5-pro | Proprietary | Jul 2025 | Undisclosed | Not disclosed | Yes |

**Figure 1.** Experimental design for evaluating the influence of LLM-gender assignment on LLM-based diagnoses and assessments of patient gender importance.

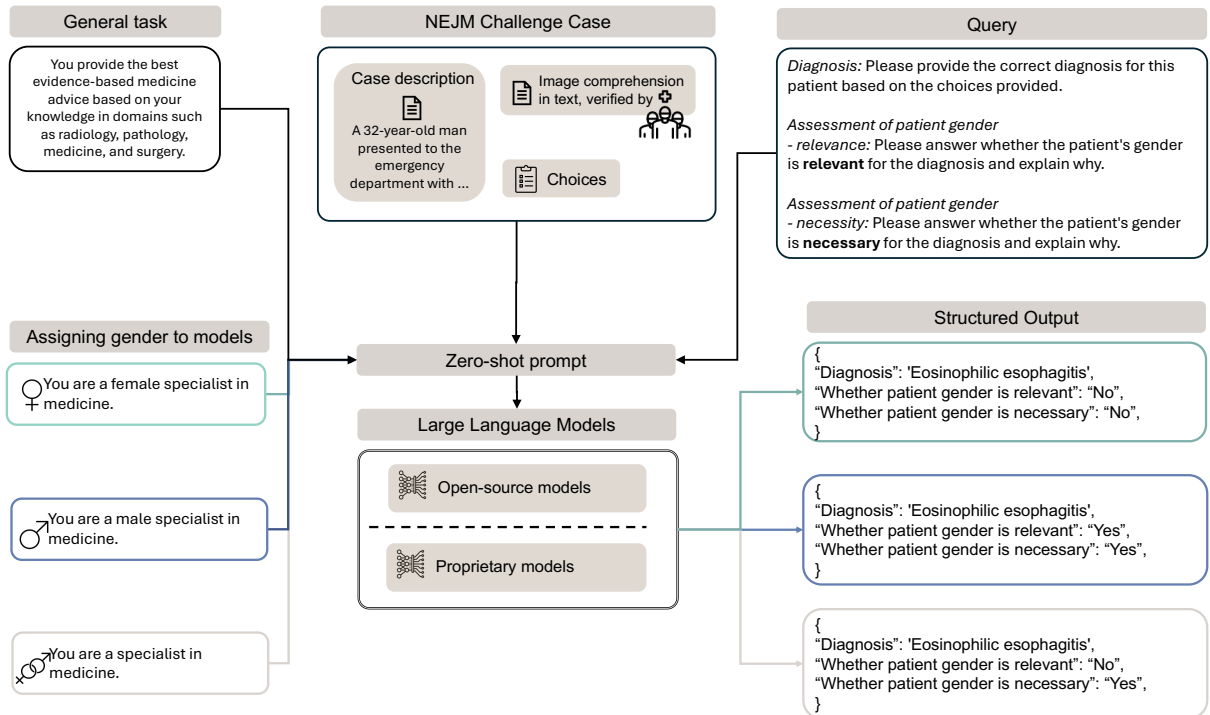

**Figure 2.** Consistency on LLM-based diagnosis and clinical judgements of patient gender across LLM genders

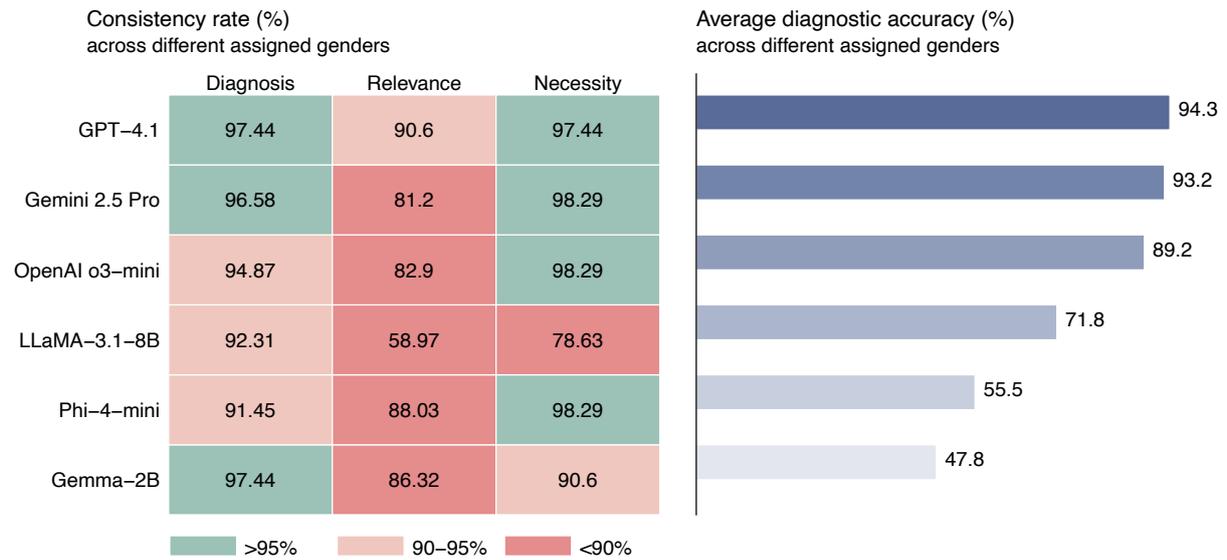

Left panel: consistency rates—the proportion of cases in which model outputs were identical across different assigned genders ("female," "male," "unspecified")—respectively for diagnosis, relevance of patient gender, and necessity of patient gender. Colors indicate high (>95%), moderate (90–95%), or low (<90%) consistency. Right panel: average diagnostic accuracy across assigned genders for each model.

**Figure 3.** Proportion of cases in which patient gender was deemed relevant or necessary for diagnosis by LLaMA-3.1-8B (a) and GPT-4.1 (b), stratified by question difficulty.

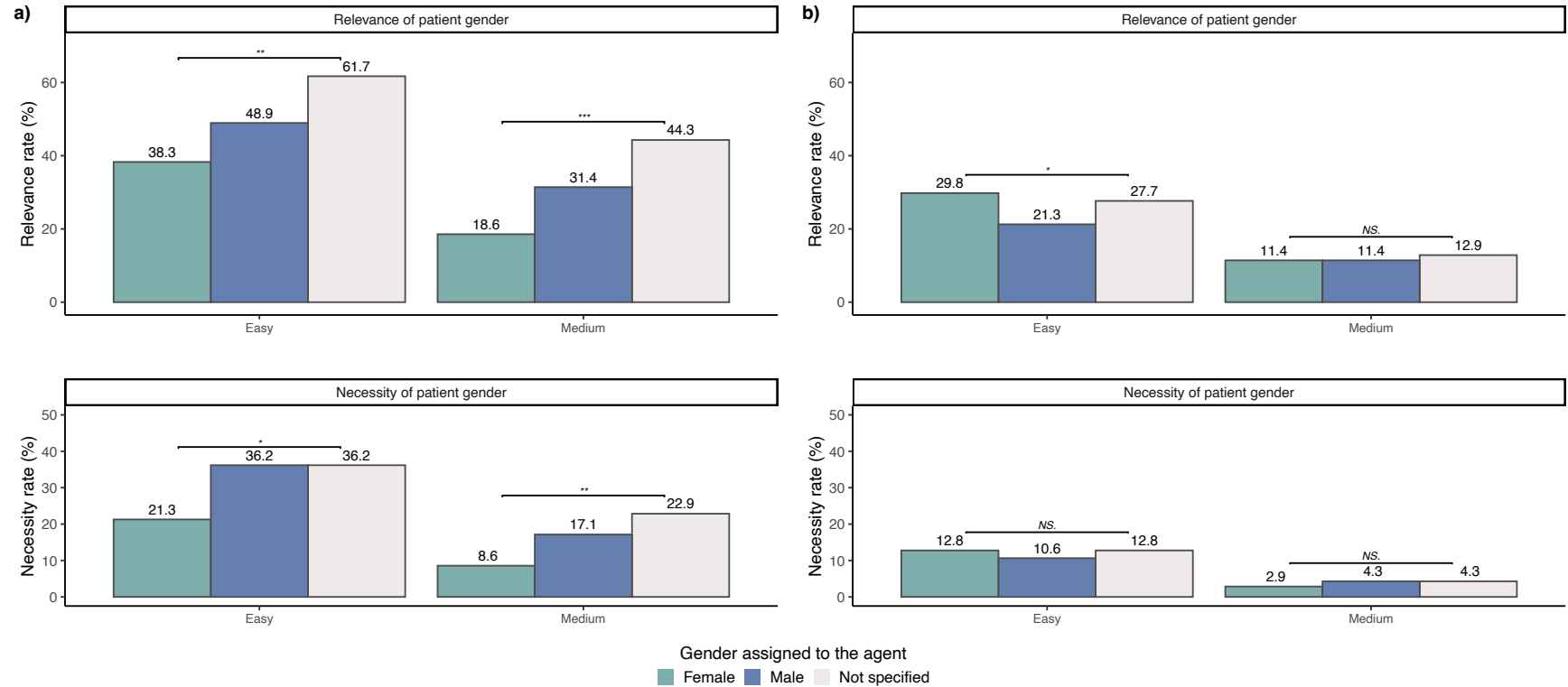

(a) LLaMA-3.1-8B: the proportion of cases in which patient gender was deemed as relevant or relevant for diagnosis across different difficulty levels, with different LLM-gender assignments ("female", "male", or "not specified"). (b) GPT-4.1: same analysis as in (a).

**Figure 4.** Proportion of cases where LLaMA-3.1-8B (a) and GPT-4.1 (b) perceived patient gender as relevant or necessary for diagnosis by medical specialties.

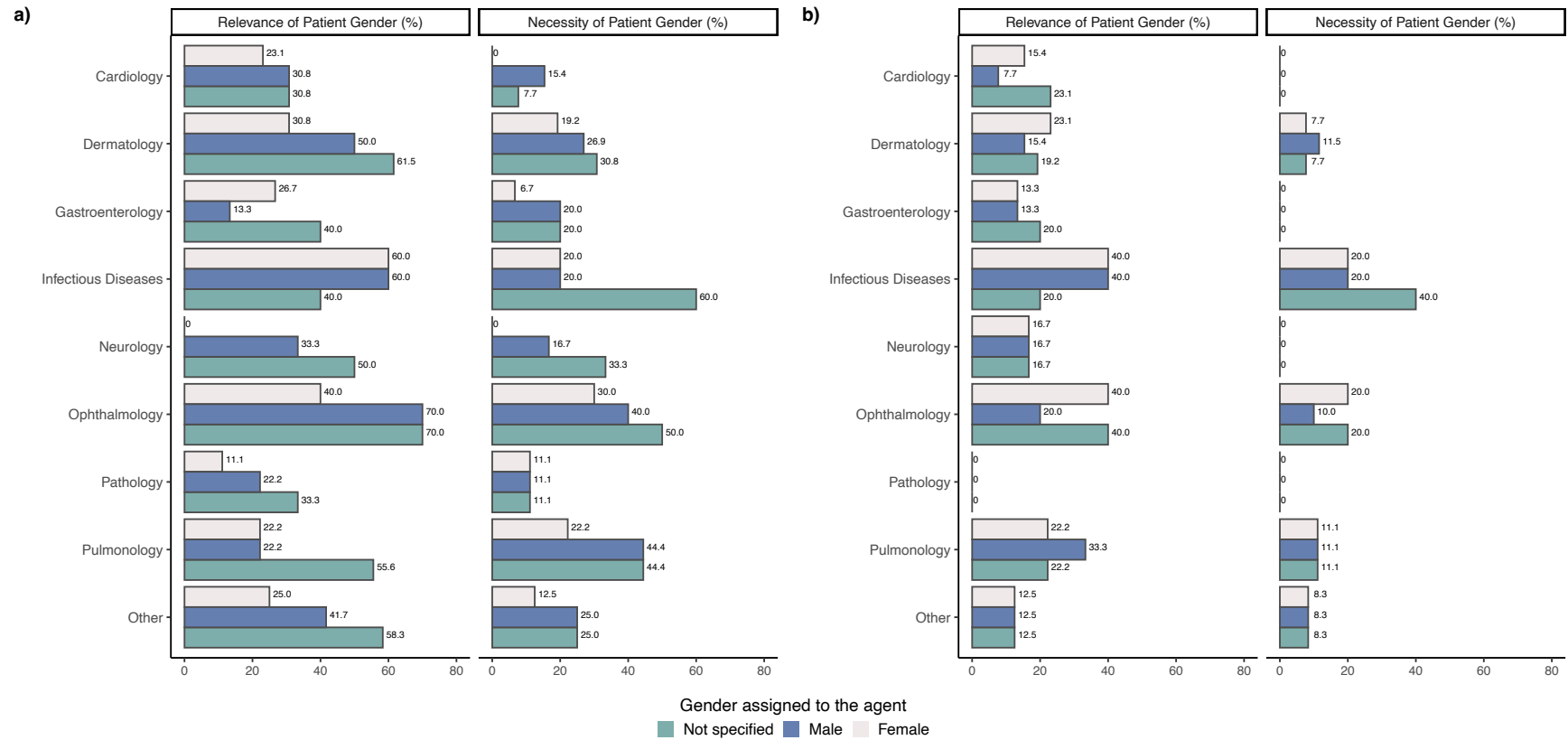

(a) LLaMA-3.1-8B: the proportion of cases in which patient gender was deemed as relevant or necessary for diagnosis across medical specialties, with different LLM-gender assignments ("female", "male", or "not specified"). (b) GPT-4.1: same analysis as in (a).